\documentclass[conference,letterpaper,10pt]{ieeeconf}
\IEEEoverridecommandlockouts
\usepackage{cite}
\usepackage{amsmath,amssymb,amsfonts}
\usepackage{algorithmic}
\usepackage{lettrine}
\usepackage{graphicx}
\usepackage{textcomp}
\usepackage{xcolor}
\usepackage{siunitx}
\usepackage{flushend}
\def\BibTeX{{\rm B\kern-.05em{\sc i\kern-.025em b}\kern-.08em
    T\kern-.1667em\lower.7ex\hbox{E}\kern-.125emX}}
\begin{document}

\title{\LARGE \bf
Momentum-Aware Trajectory Optimisation using Full-Centroidal Dynamics and Implicit Inverse Kinematics
}
\author{Aristotelis Papatheodorou, Wolfgang Merkt, Alexander L. Mitchell, Ioannis Havoutis%
\thanks{{All authors are with the Dynamic Robot Systems (DRS) group, Oxford Robotics Institute, University of Oxford. Email: \tt{\{aristotelis,wolfgang,mitch,ioannis\}@robots.ox.ac.uk}}}%
}

\maketitle

\begin{abstract}
The current state-of-the-art gradient-based optimisation frameworks are able to produce impressive dynamic manoeuvres such as linear and rotational jumps. However, these methods, which optimise over the full rigid-body dynamics of the robot, often require precise foothold locations apriori, while real-time performance is not guaranteed without elaborate regularisation and tuning of the cost function. 
In contrast, we investigate the advantages of a task-space optimisation framework, with special focus on acrobatic motions. 
Our proposed formulation exploits the system's high-order nonlinearities, such as the nonholonomy of the angular momentum, in order to produce feasible, high-acceleration manoeuvres. By leveraging the full-centroidal dynamics of the quadruped ANYmal~C and directly optimising its footholds and contact forces, the framework is capable of producing efficient motion plans with low computational overhead. Finally, we deploy our proposed framework on the ANYmal~C platform, and demonstrate its true capabilities through real-world experiments, with the successful execution of high-acceleration motions, such as linear and rotational jumps. Extensive analysis of these shows that the robot's dynamics can be exploited to surpass its hardware limitations of having a high mass and low-torque limits.
\end{abstract}

\begin{keywords}
trajectory optimisation, agile manoeuvres, full-centroidal dynamics
\end{keywords}
\vspace{-0.3cm}
\section{Introduction}\label{intro}
In recent years, the abilities of quadruped robots have improved leaps and bounds. These robots now routinely carry out autonomous inspection tasks, such as mapping construction sites to monitor progress or assist human counterparts during a maintenance schedule.
During such operations, the quadruped will encounter rugged terrains both structured and unstructured, including walking over uneven gravel surfaces and climbing stairs. 
Current commercially available controllers are capable of dealing with these scenarios both safely and reliably.
However, environments such as disaster zones or highly remote areas often exhibit terrains where agile locomotion is required to successfully operate in them.
For example, it may be necessary for the robot to jump over significant obstacles or use its momentum to stabilise over slippery or granular terrains.

\begin{figure}[t]
\centering
\includegraphics[width=3.4in]{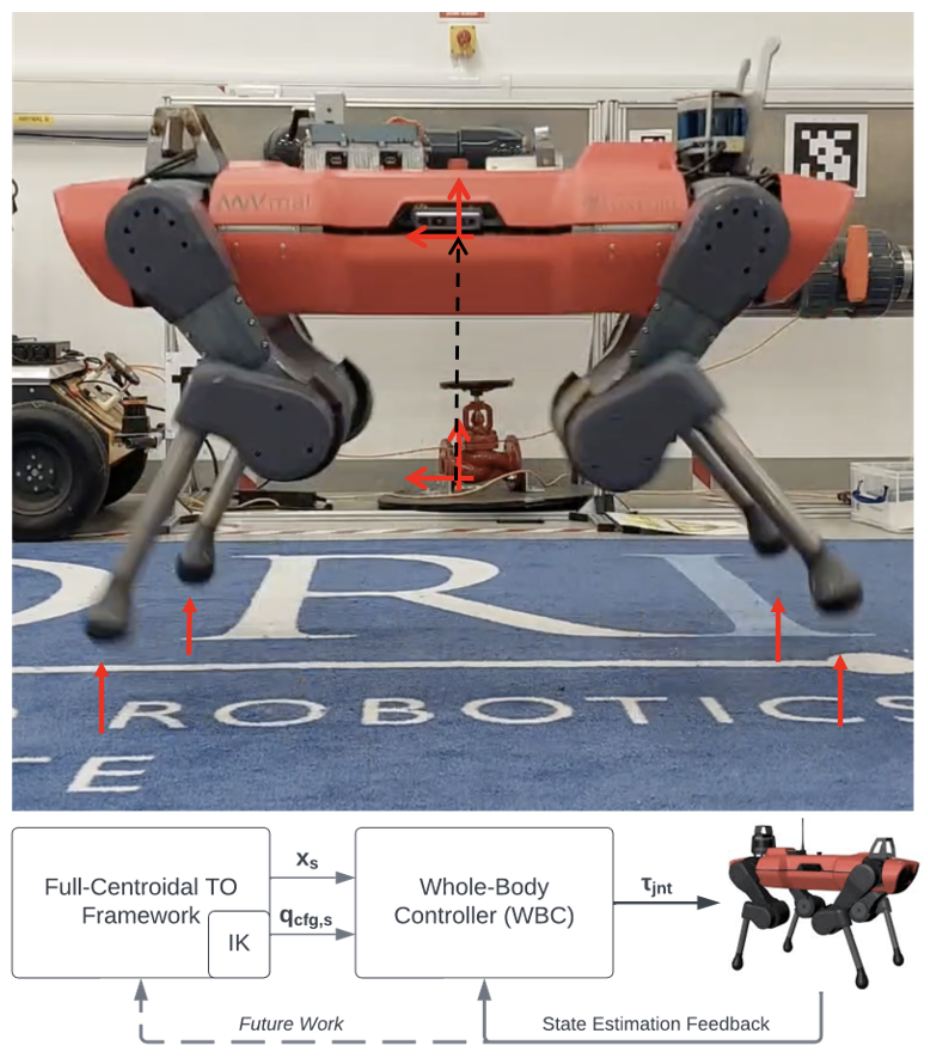}
\caption{ANYmal C performing a squat-jump: The proposed TO framework manages to overcome the limitations imposed by the high-mass of the robot and its low torque limits and successfully performs a jump. ANYmal's WBC runs at 400Hz and its not designed for acrobatic motions, hence it presents deficit tracking performance, which is partially mitigated by tuning its parameters exhaustively. The controller takes the desired state trajectory from the proposed TO framework and produces the required torques for the jump, while no re-planning is used to further highlight the ability of the proposed formulation to produce feasible, long-horizon trajectories.}
\label{architecture}
\end{figure}

Agility in legged locomotion remains an area of open research, primarily due to the complex nature of multi-contact Trajectory Optimisation (TO). The problem inherently has a combinatorial character, especially when considering articulated systems. Unlike their non-articulated counterparts, these systems have discrete contact points. The policy has to determine the sequence of these points, which in turn may not directly lead to a unique combination of limbs in contact for each timestep.
Moreover, in high-acceleration and underactuated scenarios, such as flight-phases during jumps illustrated in Fig. \ref{architecture}, kinetic momentum and particularly its angular component, imposes a nonholonomic constraint intrinsic to rigid-body systems~\cite{HolonomyNonHolonomy}.

Despite the challenges of agile legged locomotion, there are some impressive works in this area. Our approach is inspired by \cite{mastalli2022agile}, which proposes a Model Predictive Control (MPC) framework for optimising robot trajectories given control-input "box" constraints. The approach utilises a feasibility driven Differential Dynamic Programming (Box-FDDP) solver \cite{Box-FDDP} to plan and execute whole-body trajectories in an MPC fashion. Employing the full joint-space dynamics in a reactive MPC framework presents robustness and large basins of attraction for producing stable, locally optimal feedback policies. Though this model offers a temptingly high ceiling of performance, the dimensionality of the formulation may pose a significant drawback in computational performance. Furthermore, it requires precise foothold locations determined by a high-level planner or hard-coded by the user, since these cannot be discovered by the framework. Nevertheless, the authors show impressive jumping and stair climbing behaviours on a real quadruped robot.
\vspace{-0.025cm}
\subsection{Contributions}
We alternatively investigate the advantages of a \emph{task-space} formulation, which optimises the foothold locations and contact forces directly. This results in a more intuitive planning scheme, directly in the task space, capable of foothold discovery without using penalties to enforce constraints such the ones used in the full rigid-body dynamics equivalent models in \cite{mastalli2022agile} and \cite{Perceptive2024}.
In particular, a condensed full-centroidal dynamics formulation is proposed, where the state-dependent inertia and Centre of Mass (CoM) position are calculated using the closed-form Inverse Kinematics (IK) implicitly. To enhance convergence, the algorithm exploits a novel comprehensive derivation of all analytical derivatives for the robot's dynamics and cost function, while the cost's tangent-space hessian is calculated analytically without any simplifications. In doing so, we discover that it is possible to accurately control the robot’s momentum for realising agile manoeuvres such as rotational jumps that excite the high-order non-linearities of the angular dynamics.

By deploying the motions on the real ANYmal~C robot, we show that the resulting jumping trajectories map to the robot’s dynamics to such a degree that they can be tracked using a Whole-Body tracking Controller (WBC)~\cite{WBC} operating near the limits of its performance. 
This WBC incorporates a simplified dynamics model, which assumes that the robot's legs are \emph{massless}; a significant oversight for tracking the inertia of swinging limbs during agile manoeuvres.
Nevertheless, we demonstrate a range of highly-complex motions inducing aggressive evolution in the \emph{momentum} of the centroidal dynamics such as linear and rotational squat-jumps.
These manoeuvres are generated requiring minimal regularisation and we show a favourable comparison of the convergence rate of our optimisation compared to the state-of-the-art.

The proposed framework represents a significant advancement in agile locomotion. Our work extends beyond merely enhancing the original full-centroidal dynamics model with implicit foothold discovery in a compact formulation. We have also derived the fully-analytical derivatives of the dynamics and costs, leading to increased robustness in initial coniditions and high-convergence rates. To the best of our knowledge, this is the first method employing closed-form IK implicitly to reduce the dimensionality of the full rigid-body dynamics model, and show that this method can solve for the robot's contact forces and foothold locations without requiring detailed user-references apriori as regularisation.

\subsection{Related Work}\label{related}
Simplifications, such as linearised centroidal dynamics models with constant inertia, tend to be used to enforce convexity~\cite{CorberesConvexMPC, ocs2}. However, these fail to encapsulate the system's true nonlinearities, critical for dynamic stability during complex manoeuvres. The authors in~\cite{TedrakeCentroidal} represent the robot's dynamics in their centroidal-form and by using its full kinematics, they directly optimise for the joint trajectories of the actuated degrees of freedom. In comparison, our approach only optimises task-space quantities, offering a more intuitive planning scheme with lower state-space dimensionality, resulting in an efficient formulation and resolution via DDP. Reducing the dimensions of the problem is crucial, since in~\cite{mastalli2022agile} the classical full-centroidal dynamics model is proven to have the same computational complexity with the Full-Rigid Body Dynamics (FRBD) models in the quadrupeds' case. Approaches using FRBD models, such as~\cite{mastalli2023inversedynamics} \& \cite{mastalli2022agile} offer promise but suffer from their high-degree of nonlinearity along with the inherent complexities of planning in the less intuitive joint-space. It is true for \cite{mastalli2023inversedynamics} that part of the computation can be parallelised, however this requires a specialised DDP solver. Some other innovative approaches have used an alternating optimisation paradigm, attempting dynamic consensus between a reduced-order model and FRBD~\cite{MomentumAware,ADMM}. However, these approaches do not come without their set of complications. The alternating optimisation cycles, along with the inertia-dependent dynamic consensus constraints impose a performance overhead, threatening the feasibility of real-time application. Moreover, as the time-horizon increases, the growing dimensions of the Karush-Kuhn-Tucker (KKT) matrix challenge general-purpose solvers, leading to higher computational times and problematic convergence.

Differential Dynamic Programming (DDP) has emerged as a solution, excelling in performance by breaking down the problem into simpler sequences \cite{jacobson1970differential}. In particular, this family of indirect TO methods exploits the inherent markovian structure of the dynamics, avoiding the inversion of the full-KKT matrix, effectively reducing the computational complexity and large-matrix factorisations. The downside of such methods is that in their classic formulation they cannot handle constraints. Modern approaches have modified the algorithm to handle some types of constraints~\cite{Box-FDDP,ImplicitDDP} and further improve convergence, since the cost function is not filled with penalty terms. Up until now, Box-FDDP~\cite{Box-FDDP}, a state-of-the-art DDP solver with first-order Gauss-Newton approximation, handles efficiently control-input box constraints. Hence, the state-dependent dynamic-consensus constraints, which are needed by the alternating optimisation methods to achieve accurate momentum tracking, are handled as penalty terms even by the most advanced DDP solvers, adding an extra layer of complexity.

\section{Method} \label{Method}
The overall architecture of the proposed locomotion pipeline in Fig. \ref{architecture} relies on two modules: (a) the Full-Centroidal TO framework for efficiently planning trajectories, and (b) the Whole-Body-Controller (WBC) of the ANYmal~C \cite{WBC} that executes these motion plans. While creating a time-varying feedback policy based on the derived DDP gains offers certain benefits \cite{mastalli2022agile}, its inclusion is beyond the scope of this letter. The complexities of the synchronisation and real-time tuning of such a framework warrant a dedicated exploration, while they would have rendered the novelties of the proposed TO formulation insufficiently addressed.
\subsection{Problem Formulation}
The Optimal Control Problem (OCP) is illustrated in Eq. \eqref{eq:ocp_formulation} and in general requires a predefined sequence of contact points and timings. The time-horizon of the OCP is discretised in $N$ timesteps, the contributions of which are summed. The state and control-input terms are standard LQR costs. The goal is to find the optimal sequence of control-inputs $\mathbf u_s^*$ that minimises the cost and respects the given constraints. To assist convergence, we show that our formulation requires minimal regularisation in terms of $\mathbf x^{ref}$ \& $\mathbf u^{ref}$.
\vspace{-0.2cm}
\begin{equation}
\begin{aligned}
    \min _{\mathbf{u}_s \in \mathcal{U}} \sum_{k=0}^{N-1}&\left[\left\|\mathbf{x}^{ref} \ominus \mathbf{x}\right\|_{\mathbf{Q}}^2 + \left\|\mathbf{u}^{ref} - \mathbf{u}\right\|_{\mathbf{R}}^2 + l_{kin} + l_{fr}\right]_k\\
    \text{s.t.:} \quad \quad & \\
            &\mathbf x_{k+1} = \mathbf x_k \oplus \int^{t_k + dt}_{t_k}\mathbf{f\left(x_k,u_k\right)dt} \quad \text{(Dynamics)}\\
            &\mathbf{\underline u} \leq \mathbf u \leq \mathbf{\bar u} \quad \quad \quad \quad \quad \quad \text{(Control-Input Bounds)}\\
\end{aligned}
\label{eq:ocp_formulation}
\end{equation}
where:
\begin{equation}
\begin{aligned}
    \mathbf x &= \left( 
        \mathbf p_b,\; \mathbf q_o, \mathbf v_b,\; \boldsymbol \omega,\; \mathbf r_{LF},\; \mathbf r_{LH},\; \mathbf r_{RF},\; \mathbf r_{RH}
    \right) \in \mathcal X \\
    \mathbf u &= \left(\mathbf F_{LF}, \dot{ \mathbf r}_{LF}, \mathbf F_{LH}, \dot{ \mathbf r}_{LH}, \mathbf F_{RF}, \dot{ \mathbf r}_{RF}, \mathbf F_{RH}, \dot{ \mathbf r}_{RH} \right) \in \mathcal U
\end{aligned}
\label{eq:state_input_vectors}
\end{equation}

The state vector $\mathbf x$ in Eq. \eqref{eq:state_input_vectors} lies on the differentiable manifold of the admissible states ($\mathcal {X} \subseteq \mathbb{SE}(3) \times \mathbb R^{18}$) and comprises the Cartesian position and orientation of the base-origin $\left(\mathbf p_{b}, \mathbf q_{o}\right) \in \mathbb{SE}(3)$, the corresponding base velocities $\left(\mathbf v_b, \boldsymbol \omega \right) \in \mathbb R^6$ and each foothold's 3D-position for the Left-Fore ($\mathbf r_{LF}$), Left-Hind ($\mathbf r_{LH}$), Right-Fore ($\mathbf r_{RF}$) and Right-Hind ($\mathbf r_{RH}$) legs. The control-input vector $\mathbf{u}$ from Eq.\eqref{eq:state_input_vectors} resides on the Euclidean manifold of admissible controls ($\mathcal{U} \subseteq \mathbb{R}^{24}$), governed by the box constraint in Eq.\eqref{eq:ocp_formulation}. It includes each foothold's contact force ($\mathbf{F}$) and Cartesian velocity ($\dot{\mathbf{r}}$). During the swing-phase, contact forces are set to zero with free foothold velocities, whereas in the contact-phase, velocities are zero and forces are free, using the box-constraint in Eq.~\eqref{eq:ocp_formulation}. The system state, modelled as a Lie Group \cite{stillwell2008naive}, follows non-Euclidean operations like Integration $(\oplus)$ and Difference $(\ominus)$ \cite{MicroLieTheory}, excluding foothold velocities to reduce dimensionality.

The kinematic reachability term ($l_{kin,k}$) penalises the footholds that violate the robot's workspace. Essentially, a  weighted quadratic-barrier function discourages infeasible leg-configurations.
Finally the friction-cone penalty ($l_{fr,k}$) ensures that the solution does not violate the Coulomb-friction constraint for the $n_c$ feet in contact. For this, the formulation introduced in \cite{CorberesConvexMPC} is used. It constitutes an over-approximation of the friction-cone that provably has better convergence and less computational cost.
\subsection{Dynamics Model}
As previously mentioned, the Full-Centroidal Dynamics model used in this work (see Eq. \eqref{eq:cd_dynamics}) encapsulate all of the system's nonlinearities and the nonholonomic constraint of angular momentum. The chosen inertia has no simplifications, while for its calculation, the classic Articulated Body Algorithm (ABA) \cite{featherstone2014rigid} is \underline{not} used. The model has a translational part shown in Eq. \eqref{eq:trans_part} and a rotational one in Eq. \eqref{eq:rot_part}.
\begin{subequations} \label{eq:cd_dynamics}
\begin{align}
    {}^W\Dot{\mathbf{v}}_{CoM} &= \frac{1}{m}\sum^{n_c-1}_{i=0}\left({}^W\mathbf F_i \right) + {}^W \mathbf g \label{eq:trans_part}\\
    {}^W\Dot{\boldsymbol \omega} &= {}^W \mathbf I\left(\mathbf q_{cfg}\right)^{-1}[-{}^W\boldsymbol \omega \times \left({}^W \mathbf I\left(\mathbf q_{cfg}\right){}^W \boldsymbol \omega \right) \tag*{}\\
    &\quad \quad + \sum^{n_c-1}_{i=0}\left[\left({}^W \mathbf r_i -{}^W \mathbf p_{CoM} \left(\mathbf q_{cfg} \right) \right) \times {}^W\mathbf F_i \right]]
    \label{eq:rot_part}
\end{align}
\end{subequations}

The translational part is simply Newton's Second-Law, calculating the translational acceleration of the robot's CoM ($^W \Dot{\mathbf v}_{CoM}$) in which $^W \mathbf g$ represents the gravitational acceleration expressed in the World frame (W). On the other hand, the rotational part in Eq. \eqref{eq:rot_part} is not that simple. The calculation of the angular acceleration ($^W \Dot{\boldsymbol \omega}$) not only imposes the nonholonomy to the dynamics, but also major nonlinearities manifest, such as the coupling of the contact forces with the CoM and footholds' positions and the nonlinear configuration-dependent inertia matrix (${}^W \mathbf I\left(\mathbf q_{cfg}\right)$) . 

In order to calculate the inertia and CoM, the configuration of the robot's legs ($\mathbf q_{cfg}$) is necessary. However, including $\mathbf q_{cfg}$ in the state of the system would have not only been redundant, but also it would have downgraded the performance of the OCP significantly. Even though, augmenting the state with explicit joint-angles, as opposed to their implicit calculation, renders the problem more sparse, this does not mitigate the inherent nonlinearities and the associated local minima of the underlying OC problem, as the dynamics continue to be represented by the same model in both scenarios. Furthermore, such augmentation necessitates an additional four tensor contractions for the computation of the dynamics' derivatives, attributed to the requirement of differentiating the \textit{inverse} kinematic Jacobians that map the foothold velocities to the joint ones, as Eq.~\eqref{eq:diff_kin} illustrates.
\begin{equation}
    \dot{\mathbf q}_{cfg} = \mathbf J^{-1}\left(\mathbf q_{cfg}\right)\dot{\mathbf r}
\label{eq:diff_kin}
\end{equation}

By using the closed-form IK implicitly, the aforementioned shortcomings are effectively handled, while the requirement of having the analytic IK solution does not restrict our framework, as most quadrupeds have three-joint legs. According to Eq. \eqref{eq:IK}, given the current state of the system, the algorithm can produce the joint angles of the robot's legs.
\begin{equation}
    \mathbf q_{cfg,k} = \mathbf{IK}\left( \mathbf x_k\right), \; \text{with:} \; k \in \left[0, N \right)
    \label{eq:IK}
\end{equation}

Ultimately, the dynamics constraint of the OCP in Eq. \eqref{eq:ocp_formulation} is composed as in Eq. \eqref{eq:dynamics} and lives in the Euclidean tangent-space ($\mathcal{T_X}$) of the admissible states. More specifically, the control-input directly affects the footholds, which in turn, along with the contact forces, affect the CoM's dynamics (see Eq. \eqref{eq:cd_dynamics}).
\begin{equation}
\begin{aligned}
    &\Dot{ \mathbf x} =\mathbf{f(x,u)}=\\
    &\begin{pmatrix} ^B \mathbf v_{b} & ^B \boldsymbol \omega & ^B \dot{\mathbf v}_{b} & ^B \Dot{\boldsymbol \omega} & ^W\dot{\mathbf r}_{LF} & \hdots {}^W\dot{\mathbf r}_{RH} \end{pmatrix} \in \mathcal{T_X} \subseteq \mathbb{R}^{24}
\end{aligned}
\label{eq:dynamics}
\end{equation}
However, the translational acceleration of the CoM has to be projected to the base-origin of the robot ($^B \Dot{ \mathbf v}_b$) expressed in its body-fixed frame (B). The formula in Eq. \eqref{eq:base_acc} approximates this projection by neglecting minor gyroscopic terms, given that the relative acceleration and velocity of the CoM to the robot's base-origin remains low. The alternative would be to employ \emph{Pinocchio's} ABA algorithm to calculate the missing terms, but that comes with significant computational overhead.
\begin{equation}
\begin{aligned}
    {}^B \dot{\mathbf v}_{b} &= \;^B \mathbf R_W{}^W \Dot{\mathbf v}_{CoM} + \left({}^B \mathbf R_W{}^W\Dot{\boldsymbol\omega} \right) \times {}^B\mathbf p_{CoM \rightarrow b}\\
    &\quad \quad+{}^B\boldsymbol \omega \times \left(^B \boldsymbol \omega \times {}^B \mathbf p_{CoM\rightarrow b}\right) - \;^B\boldsymbol \omega \times {}^B\mathbf v_b
\end{aligned}
\label{eq:base_acc}
\end{equation}
\subsection{Calculation of Derivatives}
Box-FDDP is a gradient-based method. Hence, the state and control-input derivatives of the dynamics and cost, along with the cost's hessians are required. The solver approximates the cost with a linear quadratic function and finds the descend direction at each iteration. The inherent limitation of vanilla DDP solvers is that they cannot handle state-dependent constraints explicitly, so kinematic reachability and friction-cone constraints must be encoded as penalties, adding complexity and sensitivity to the cost function. In order to have robust convergence and increase the overall performance of the algorithm, an efficient derivation of the fully analytical derivatives and cost-hessians has been performed. This proved to be a challenging task.

Our hybrid state-vector excludes foothold velocities, requiring the implementation of a custom \emph{Crocoddyl} action model. For integration, the symplectic Euler method from the library was used, a variational integrator \cite{west_variational_2004} that preserves the system's geometric properties and phase-space symmetries, ensuring better energy conservation.

Additionally, the state derivatives reside in the tangent space $\mathcal{T_X}$. Most tangent-space first-order derivatives (e.g. for the Integration and Difference operators) are well documented in the literature \cite{MicroLieTheory}, while \emph{Pinocchio} provides efficient implementations. However, the state cost-term requires the hessian of the Difference operator in order to come with the analytic hessian of the overall cost function. None of the available frameworks and publications \cite{MicroLieTheory} \cite{Pinocchio} provide an implementation, while its effect on convergence has not been explored before. To start with, the first-order derivative of the state cost-term is calculated as shown in Eq. \eqref{eq:l_state_x}, where $\otimes$ designates the Hamilton product \cite{Jia2015QuaternionsAR} and $\mathcal{D}$ the euclidean tangent-space differentation operator.
\begin{equation}
\begin{aligned}
    &l_{\mathbf x} = \left(\mathbf{x^{ref}} \ominus \mathbf{x} \right)^\top \mathbf Q \frac{\mathcal{D} \ominus}{\mathcal{D} \mathbf x}\;,\\
    \text{where:}\quad  \frac{\mathcal{D} \ominus}{\mathcal{D} \mathbf{x_{SE3}}} &= - \mathbf{J^{-1}_{l,SE3}}\left( \boldsymbol \tau \right),\\
    \text{with:}\ \boldsymbol \tau = \left(\mathbf p, \boldsymbol \theta \right) &= \mathbf{x^{ref}_{SE3}} \ominus \mathbf{x_{SE3}} = \mathbf{Log}\left( \mathbf{x^{-1}_{SE3}} \otimes \mathbf{x^{ref}_{SE3}} \right)
\label{eq:l_state_x}
\end{aligned}
\end{equation}

Note that the derivative of the Difference operator is non-trivial only for the $\mathbb{SE}(3)$ elements of the state (i.e. $\mathbf{x_{SE3}}$), hence Eq. \eqref{eq:diff_x} shows the derivation only for these elements. Note that $\left(\mathbf p, \boldsymbol \theta \right)$ represent the tangent-space lie-algebra vectors of the manifold \cite{MicroLieTheory}. These vectors are related to the base's position and orientation with the Exponential and Logarithmic mappings. Furthermore, the left-Jacobian ($\mathbf{J_{l,q_o}}$) of the quaternion ($\mathbf q_o$), along with matrix $\mathbf Q$ are derived in \cite{MicroLieTheory}.
\begin{equation}
\begin{aligned}   
    \mathbf{J^{-1}_{l,SE3}}\left( \boldsymbol \tau \right) =
    \begin{pmatrix}
    \mathbf{J^{-1}_{l,q_o}}\left( \boldsymbol \theta \right) && \mathbf{J^{-1}_{l,q_o}}\left( \boldsymbol \theta \right) \mathbf{Q\left(\boldsymbol \tau \right)} \mathbf{J^{-1}_{l,q_o}}\left( \boldsymbol \theta \right) \\
    \mathbf 0 && \mathbf{J^{-1}_{l,q_o}}\left( \boldsymbol \theta \right)
    \end{pmatrix}\\
\label{eq:diff_x}
\end{aligned}
\end{equation}

\begin{figure*}[t]
\vspace{0.3cm}
\centering
\includegraphics[width=1\linewidth]{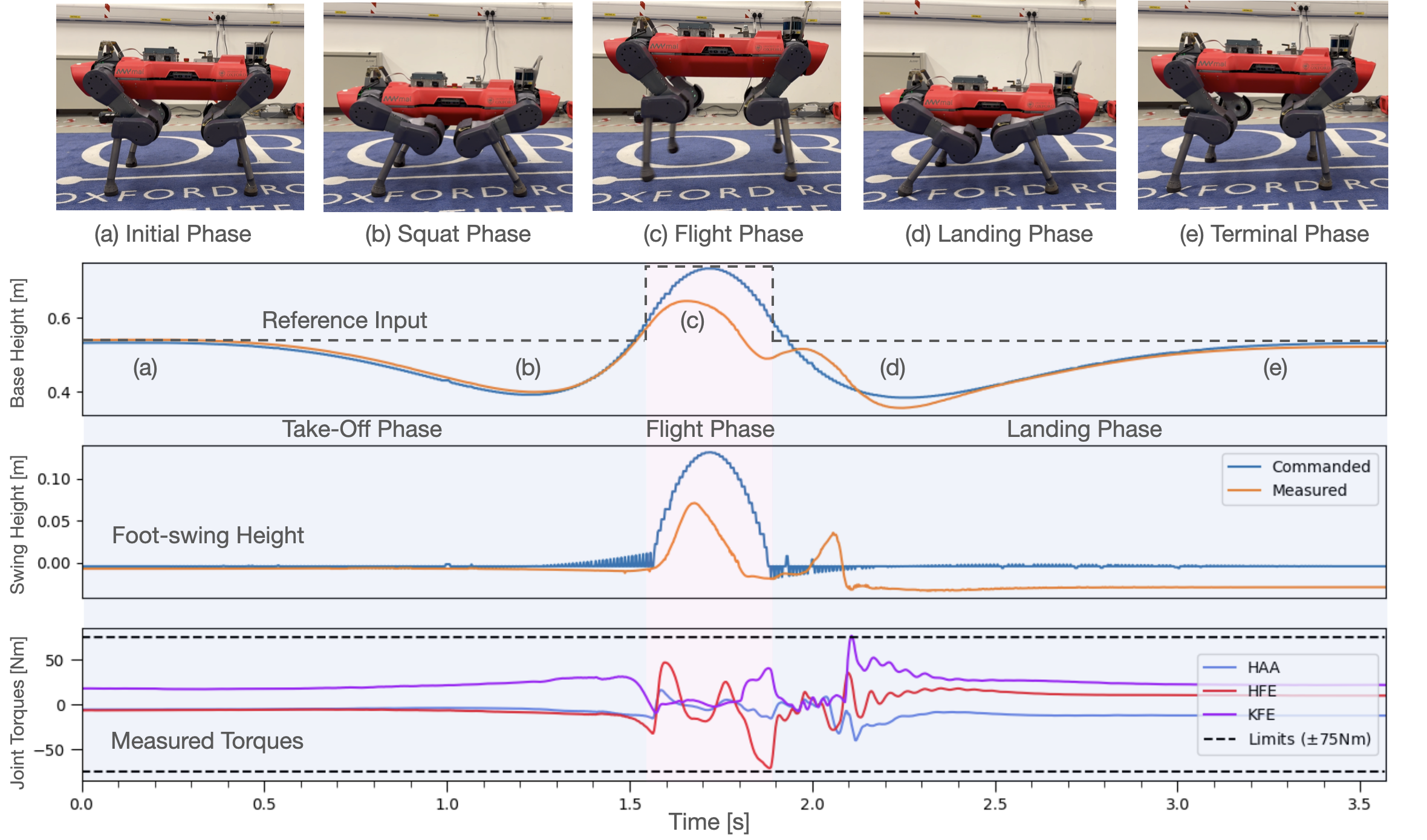} 
\caption{A linear squat-jump as deployed on the real ANYmal~C robot. The top row shows the robot manoeuvre during five distinct phases: (a) the initial condition, (b) the squat phase, (c) in-flight phase, (d) landing period, and (e) return to initial conditions. The snapshots of the robot found in the first row are accompanied by the trajectories for the base height (row 2), foot-swing (row 3) and the measured joint torques (row 4). The input regularisation to the TO is the desired reference height as indicated in the base-height plot, while no references are required for the swing trajectories. The measured torques during the manoeuvre do not violate the torque limits of \SI{75}{\newton\meter} even during touchdown. The negative foot height after touchdown is attributed to state-estimation drift, as the state estimator is not designed to handle flight phases with sudden impacts effectively.}
\label{fig:linear_jump_plot}
\end{figure*}
By differentiating Eq. \eqref{eq:l_state_x} again, the hessian of the state-cost term is derived in Eq. \eqref{eq:l_state_xx}. Even though the Difference operator's hessian is a tensor, its sparsity is leveraged to come with an efficient implementation. Experimentation has shown that the inclusion of this hessian increases the solver's robustness to infeasible starts. Further details on the resolved tasks can be found in Section \ref{Results}.
\begin{equation}
\begin{aligned} 
    &l_{\mathbf{xx}} = \frac{\mathcal D \ominus}{\mathcal D \mathbf x}^\top \mathbf Q \frac{\mathcal D \ominus}{\mathcal D \mathbf x} + \left(\mathbf{x^{ref}} \ominus \mathbf{x} \right)^\top \mathbf Q \frac{\mathcal D ^2 \ominus}{\mathcal D \mathbf x^2} \\
    &\frac{\mathcal D ^2 \ominus}{\mathcal D \mathbf x^2_{\mathbf{SE3}}} = \frac{\partial}{\partial \boldsymbol \tau}\left( - \mathbf{J^{-1}_{l,SE3}}\left( \boldsymbol \tau \right) \right) \frac{\mathcal D \ominus}{\mathcal D \mathbf{x_{SE3}}}    
\end{aligned}
\label{eq:l_state_xx}
\end{equation}

Finally, the inertia matrix's derivatives posed a significant challenge. In any composite rigid-body system, the total inertia can be calculated by sequentially projecting each body's inertia to a specified frame (e.g. body-fixed frame) and accumulating their contributions with a weighted sum. Hence, the derivative of the system's inertia is retrieved by differentiating these projections w.r.t. the legs' configuration ($\mathbf q_{cfg}$) and summing them. However, by exploiting the articulated nature of the system, these computations can be parallelised to calculate each leg's inertia simultaneously. Ultimately, to come up with the derivatives w.r.t. the system's state $\mathbf x$, the closed form IK are auto-differentiated using CasADi \cite{CasADi}. Lastly, the tangent-space derivative of the inertia are computed using the chain-rule as shown in Eq.~\eqref{eq:inertia_x}. Once again, their sparsity is leveraged to optimise their runtime.
\begin{equation}
     \frac{\mathcal D ^B\mathbf I}{\mathcal D \mathbf{x}} = \frac{\operatorname{d} {}^B\mathbf I}{\operatorname{d} \mathbf{q}_{cfg}}\frac{\mathcal D\mathbf{q}_{cfg}}{\mathcal D \mathbf x} 
\label{eq:inertia_x}
\end{equation}

A workspace violation avoidance mechanism is needed to prevent kinematic singularities and irregular configurations. Footholds violating kinematic constraints are normalised within the leg's effective workspace before being passed to the IK algorithm.

\section{Results \& Discussion} \label{Results}

\begin{figure*}[t]
\centering
\includegraphics[width=1.04\linewidth]{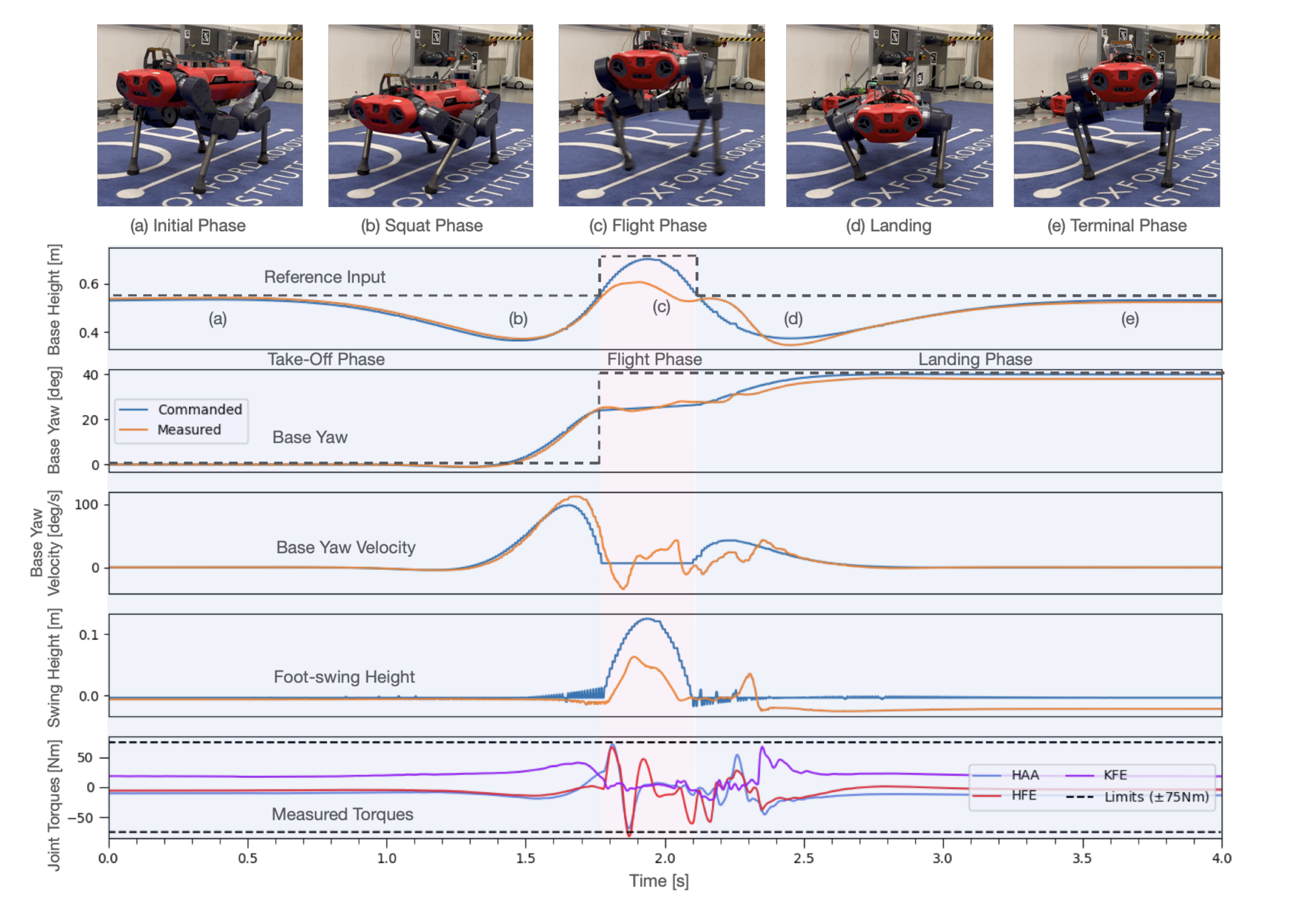}
\vspace{-0.9cm}
\caption{A \SI{40}{\deg} rotational jump deployed on the real robot. Only the desired yaw angle and base-height are provided to the solver, along with an approximate regularisation for the feet locations at touchdown. The top row of this figure shows five snapshots during the agile manoeuvre: (a) the initial condition, (b) squat phase, (c) flight phase, (d) landing phase and (e) return to initial conditions. The second and third rows show the base height and yaw trajectories. The fourth row shows the foot-swing tracking, while the final row shows the measured torques for the front left leg which evolve smoothly and do not violate the torque limits of \SI{75}{\newton\meter}. Note the smooth evolution of the angular velocity before and during the flight-phase, indicating the good inertia tracking capabilities of the controller. The yaw velocity is brought close to zero during the flight-phase in order to land smoothly and not violate the friction-cone constraint at touchdown.The negative foot-height after touchdown is attributed to state-estimation drift, as the state estimator is not designed to handle flight phases with sudden impacts effectively.}
\label{fig:rotational_jump}
\vspace{-0.45cm}
\end{figure*}

Analysis of the capabilities and opportunities of the approach is guided by the following questions: (A) Can this task-space framework  produce feasible long-horizon trajectories for agile manoeuvres? 
(B) To what degree does this formulation control the robot's momentum and how the tracking capabilities of the simplified Whole-Body Controller (WBC) affect the robot's motion? (C) Can the approach generate feasible robot trajectories given minimal reference inputs? 
(D) How does the convergence speed of the centroidal dynamics model compare to the state-of-the-art?

In the following sections, we investigate these questions by analysing the results from our real-world experiments in order to shed light on the TO framework's performance, tuning process and potential limitations.

\subsection{Optimising Long-Horizon Trajectories}

Three manoeuvres are optimised using our task-space framework.
The executed motions are presented in increasing relative agility in order to highlight different aspects of the framework's behaviour. We first present a lemniscate-tracking task (figure-of-eight motion), before we progress to linear \& rotational squat-jumps.
The final two manoeuvres can only be successfully executed if the robot's base and leg inertias are modelled and controlled with a high degree of fidelity and thus these manoeuvres are real tests of our task-space optimisation framework.

To tune the tasks, the Weights \& Biases (wandb) \cite{wandb} hyperparameter optimisation package was used. Its Bayesian optimisation module was utilised to find a close-to-optimal solution for the weights, while the kinematic reachability and Coulomb friction constraints where fine-tuned manually afterwards for best performance. Arguably by defining a more elaborate minimisation function, the manual fine-tuning step could be eliminated entirely. 
\subsubsection{Lemniscate Tracking}
Even though it seems trivial, the "8-like" motion of the base highlights the ability of the planned trajectory to respect the imposed penalty-constraints, since the CoM is positioned near the margins of the legs' stability polygon without violating kinematic or friction-cone constraints. Non momentum-aware controllers could not realise such trajectory, since they assume fixed CoM w.r.t. the base-origin, hence they do not have accurate control authority over its actual position. The task is presented in the accompanying video in the interest of space.

\subsubsection{Squat Jump}
The squat-jump is designed to stress ANYmal's WBC to the limit of its tracking capabilities. This proved to be very demanding, since the robot has \emph{significantly higher mass} (\SI{55}{\kilo\gram}) and more \emph{restrictive joint-torque limits} (\SI{75}{\newton \meter}), compared to the ones that have demonstrated similar behaviours in the past \cite{CorberesConvexMPC, MomentumAware}. The task consists of an abrupt jumping motion that requires the solver to plan a deep-squat trajectory to gain the required momentum, since the robot's low power density is restrictive. To maintain stability, the legs have to be moved accordingly in order to sustain correct orientation and base-height recovery during the landing phase. 
The duration of the task is \SI{6.03}{\second} and the base height, footswing and torque tracking results from the real-world experiment are illustrated in Fig.~\ref{fig:linear_jump_plot}. Note that the duration of the flight-phase was tuned using wandb, while the take-off and landing phases were manually defined to achieve smooth tracking.

\subsubsection{Rotational Jump}
This manoeuvre requires careful control of the robot's angular momentum during both contact and flight phases of the motion.
During the motion, which is presented in Fig.~\ref{fig:rotational_jump}, the method generates angular momentum during the squat phase (Fig~\ref{fig:rotational_jump}~(b)) and regulates it during the flight phase (Fig~\ref{fig:rotational_jump}~(c)). This angular momentum can only be tracked through the secondary effects resulting from the motion of the swinging legs and the configuration-dependent inertia and CoM. This is a very complex manoeuvre and highlights the benefit of our formulation, while the simplified WBC's tracking is sufficient given its simplifications.

\begin{table*}[t]
\vspace{0.3cm}
    \centering
    \caption{Minimal references are required to generate complete task-space trajectories for squat-jump and rotational-jump. Phases (a), (c), and (e) correspond to those in Fig.~\ref{fig:linear_jump_plot} for the stand, flight, and landing phases, respectively}
    \label{tab:I_O_table}
    \begin{tabular}{c|ccc|ccc}
        \multicolumn{1}{c}{} & \multicolumn{3}{c}{Reference Inputs} & \multicolumn{3}{c}{TO Outputs (Supplied to the WBC)} \\
        Manoeuvre Phase  & Base Height [m] & Base Yaw [deg] & Swing Height [cm] &  Base Height [m] & Base Yaw [deg] & Swing Height [cm] \\
         \hline \\
        Squat Jump (a)   & 0.52 & Not Req'd & Not Req'd & 0.52 & 0.0 & 0.0  \\
        Squat Jump (c)   & 0.72 & Not Req'd & Not Req'd & 0.73 & 0.0 &  13.1 \\
        Squat Jump (e)   & 0.52 & Not Req'd & Not Req'd & 0.52 & 0.0 &   -0.3\\
        Rot Jump (a)     & 0.52 & 0.0       & Not Req'd & 0.52 & 0.0 &   0.0 \\
        Rot Jump (c)     & 0.70 & 40.0      & Not Req'd & 0.61 & 40.0 & 12.5  \\
        Rot Jump (e)     & 0.52 & 40.0      & Not Req'd & 0.52 & 40.0 & -0.4  \\
    \end{tabular}
    \vspace{-0.5cm}
\end{table*}

\subsection{Momentum Awareness \& WBC Tracking}

The TO framework's ability to create realisable trajectories is analysed here.
The produced trajectories created by our TO framework are tracked using the robot's WBC.
The WBC is constrained using a reduced-order model, with the crucial assumption of \emph{massless} limbs. In this section we investigate the capabilities of our framework in terms of controlling the robot's momentum and how the simplifications of the WBC affect the end-result. 

Fig.~\ref{fig:linear_jump_plot} shows the tracking performance for the linear jump.
The behaviour can be broken down into three distinct phases: the initial squat, the in-flight phase, and the landing. The TO framework is able to plan the jumping motion, by performing a squat to build-up the required momentum, and by moving the limbs during flight-phase to balance the base and have a smooth touchdown. 
Moreover, the WBC is able to track the initial and final phases well, when all four feet are in contact. However, during the most agile flight-phase, the tracking diverges. 
This is expected as the WBC cannot build-up adequate momentum, since its simplifications create sub-optimal feed-forward torques. Our results indicate that the joint torques during take-off are kept relatively low (except at the transition), which means that the robot could jump higher and reach the desired height using a momentum-aware torque controller. Note that the WBC tracks the trajectory with fixed gains at each jumping phase, hence increasing its tracking PD gains would lead to major torque violations during the contact transitions. Moreover, the robot is not equipped with force sensors at its feet, hence the state-estimator indirectly predicts contacts using the robot's kino-dynamic model. In abrupt, high-acceleration flight phases, faulty estimation results in \emph{ringing} contact feedback, which destabilises the controller and results in the out-of-plan double-stepping. Note that we have already tuned the simplified WBC to extract performance, but it is impossible to track momentum adequately to mitigate these issues, hence further investigation towards this seems futile. It's true that with a better tracking controller (e.g. \cite{mastalli2022agile}), our framework could yield the optimal results, but even now the experimental validation is impressive in terms of feasibility and momentum regulation.

The TO is able to use the robot's momentum to perform rotational jumps in a controllable fashion and this capability is analysed here.
In the rotational jump's case in Fig. \ref{fig:rotational_jump}, our TO framework increases the base's yaw velocity before take-off and smoothly decelerates to a very low angular velocity. At touchdown, the framework induces additional rotational acceleration to smoothly position the base in the desired \SI{40}{\degree} orientation. The desired yaw velocity profile in the transition between flight and contact phases is abrupt and this exposes one limitation of the full-centroidal dynamics formulations in general, which we further analyse in Section~\ref{limitations}. Nevertheless, WBC's tracking error is small, and the overall system's response is smooth. This is partially due to the accurate inertia tracking capabilities of our method, since by accelerating the base proactively during contact and transitioning to a smooth flight-phase, no friction or torque constraints are violated during the contact transitions.

\subsection{Trajectory Optimisation with Minimal Reference Inputs}

The ability to generate feasible and acrobatic motions with minimal regularisation is investigated in this section. As discussed in Section \ref{intro}, most optimal control methods tend to require detailed references ($\mathbf x^{ref},\; \mathbf u^{ref}$ in Eq.~\eqref{eq:ocp_formulation}) by the user, in addition to the hard-coded contact timings and foothold locations. During the experiments, it is shown that our TO method only requires minimal regularisation. This is summarised in Table~\ref{tab:I_O_table} for both the linear \& rotational jumps. In particular, the linear squat-jump requires only the desired base-height as regularisation, whilst the rotational-jump requires the desired base-yaw and rough touchdown foothold locations in addition to the base-height. In order to reach the desired base-height during the squat-jump, the TO method produces a squatting motion and naturally drops its base before accelerating into a jump. In doing so, the method optimises the feet swing-heights given predefined timings, without needing desired swing-heights as regularisation for all jump types. In conclusion, our framework produces feasible trajectories for both the base and the feet using only minimal user-references.
\vspace{-0.1cm}
\begin{table}[h]
    \centering
    \caption{Achieved single-thread performance for the dynamics' integration (\emph{calc}) and derivatives (\emph{calcDiff}) for a single time-step (*Different Task-Setup).}
    \label{tab:Perf_table}
    \begin{tabular}{c|ccc|ccc}
        \multicolumn{1}{c}{} & \multicolumn{2}{c}{Dynamics [$\mu$s]} & \multicolumn{2}{c}{Convergence [$iter.$]} \\
        Model & calc  & calcDiff & Jump & Rot. Jump\\
         \hline \\
        FRBD \cite{Box-FDDP} ($\pm \sigma$) & 11.48$\pm$3.94 & \textbf{29.11$\pm$10.51} & 53* & -\\
        Ours ($\pm \sigma$) & \textbf{7.68$\pm$2.96} & 48.14$\pm$9.33 & \textbf{24} & \textbf{27}
    \end{tabular}
    \vspace{-0.3cm}
\end{table}

\subsection{Computational Performance}

Our condensed formulation results in a a lower-dimensionality than the FRBD equivalent models and the improvements in convergence speed  are analysed here. According to Table \ref{tab:Perf_table}, our framework achieves a considerable speedup not only in the integration of the dynamics, but also in the demonstrated convergence compared to methods in literature \cite{Box-FDDP}. Finally, our code-base is not optimised for real-time performance (hence the higher runtime of our method's \emph{calcDiff()} compared to the FRBD's one), since no re-planning is used in the current work, thus its true potential is expected to be higher. Both tasks in Table \ref{tab:Perf_table} have 6.03s duration without warm-starting, while the runtime statistics are calculated for 1000~samples. Note that planning such long-horizon trajectories with penalty-based solvers, such as Box-FDDP, is challenging since soft-constraints result in local-minima, compromising convergence.

\section{Limitations \& Future Work}\label{limitations}
The rotational jump scenario exposes a key limitation of the full-centroidal dynamics model; its momentum tracking capabilities are limited due to its single, composite rigid-body assumption. Unlike its FRBD counterpart, the composition of multiple rigid-bodies into a single one does not account for the momentum changes of each individual link. An illustrative example is the \emph{falling cat} re-orientation problem, where a cat in free fall can reorient its body without an initial rotation \cite{HolonomyNonHolonomy}. Even though our current hardware cannot imitate such behaviours, FRBD models are in theory capable of capturing such effects. On the other hand, by reviewing Eq.~\eqref{eq:rot_part}, it becomes clear that the full-centroidal dynamics model is not able to change its orientation, starting from zero-initial conditions, without the application of an external force. Hence, these models are incapable of precise orientation tracking in extensive flight-phases, since the momentum-changes of the individuals limbs are neglected \cite{orin_centroidal_2013}. This limitation is not prohibitive, since it occurs when the relative to body accelerations of the limbs become significant, effectively amplifying their inertial effects. Our results prove that for short and agile flight-phase motions, these phenomena can be neglected up to a certain extent, while the performance advantages of having a simpler formulation, outweigh the fidelity of a more complex model in most quadrupedal use-cases. Note that this limitation should not be confused with the double-stepping effect caused by the deficit in tracking performance of the WBC.

Substituting the WBC will be a key area to focus our future research efforts. A promising direction would be to evaluate the framework's real-time capabilities and create an optimised MPC that bypasses ANYmal's simplified controller by using the locally optimal state-feedback policies derived from the Box-FDDP solver's solution. However, the local-approximation nature may have destabilising effect on the real-robot, given that the task-space quantities that we are solving for need to be translated to joint torques using noisy state-estimation. Another path could be the adoption of an MPC such as in~\cite{mastalli2022agile}. Our framework can be used to produce long-horizon, acrobatic motions and warm-start the FRBD MPC that can track these trajectories and avoid the shortcomings of the single rigid-body assumption of the full-centroidal dynamics model, resulting in improved tracking.

\section{Conclusion} \label{Conclusion}
The quest for enhanced agility in legged robots, particularly within the realm of trajectory optimisation, remains a challenging endeavour. This work presented a comprehensive trajectory optimisation framework tailored for the quadruped ANYmal~C. We introduced a superior Full-Centroidal Dynamics formulation, outperforming existing methods\cite{CorberesConvexMPC, TedrakeCentroidal, MomentumAware}, with automatic foothold discovery and precise inertia tracking, without redundant states. To the best of our knowledge, this is the first to implicitly incorporate analytic IK for improved performance without compromising model fidelity. Our innovations enhance convergence and simplify planning by optimising task-space quantities with minimal regularisation, validated through real-world experiments, as detailed in Section \ref{Results}.

\bibliographystyle{IEEEtran} %
\bibliography{references} %
\end{document}